\pdfoutput=1
\documentclass[11pt]{article}

\usepackage{EMNLP2023}

\usepackage{times}
\usepackage{latexsym}

\usepackage[T1]{fontenc}
\usepackage[utf8]{inputenc}

\usepackage{microtype}

\usepackage{inconsolata}
\usepackage{tabularx}

\usepackage{colortbl}

\title{Instructions for EMNLP 2023 Proceedings}

\author{First Author \\
  Affiliation / Address line 1 \\
  Affiliation / Address line 2 \\
  Affiliation / Address line 3 \\
  \texttt{email@domain} \\\And
  Second Author \\
  Affiliation / Address line 1 \\
  Affiliation / Address line 2 \\
  Affiliation / Address line 3 \\
  \texttt{email@domain} \\}

\usepackage{inconsolata}
\usepackage{url}
\usepackage{linguex}
\usepackage{babel}
\usepackage{natbib}
\usepackage[useregional]{datetime2}
\usepackage{graphicx} 
\usepackage{amsmath}
\usepackage{amsfonts}
\usepackage{amssymb}
 \usepackage{mathtools} 
\usepackage{caption}
\usepackage{subcaption}
\usepackage{dsfont}
\usepackage{mathtools}
\usepackage{bbm}
\usepackage{bm}
\usepackage{multirow}
\usepackage{multicol}
\usepackage{natbib}
\usepackage{tikz}
\usepackage{booktabs}
\usepackage{adjustbox}
\usepackage{amssymb,amsmath,linguex,amsthm}
\theoremstyle{definition}

\newcommand{\hidden}[1]{}

\title{Nebula: A Discourse-Aware Minecraft Builder}
\author{Akshay Chaturvedi$^{\dagger}$, Kate Thompson*$^{\dagger}$, Nicholas Asher$^{\dagger\ddagger}$\\ 
       $^{\dagger}$IRIT, $^{\ddagger}$CNRS, *LINAGORA Labs \\ Toulouse, France }
\date{\today}

\begin{document}

\maketitle
\begin{abstract}
When engaging in collaborative tasks, humans efficiently exploit the semantic structure of a conversation to optimize verbal and nonverbal interactions.  But in recent ``language to code'' or ``language to action'' models, this information is lacking.  We show how incorporating the prior discourse and nonlinguistic context of a conversation situated in a nonlinguistic environment can improve the ``language to action'' component of such interactions.  We finetune an LLM to predict actions based on prior context; our model, Nebula, doubles the net-action F1 score over the baseline on this task of \citet{jayannavar:etal:2020}.   We also investigate our model's ability to construct shapes and understand location descriptions using a synthetic dataset.
\end{abstract}

\section{Introduction}

High level building agents use conversation in a collaborative task to combine information about the extant conversation, the world, and prior actions to execute new instructions. Such agents interpret messy or vague language, produce actions, then reassess the situation, ask questions or take in corrections from other agents to optimize their actions. Successful collaborative conversations are vital for efficiently performing complex interactive tasks. In this paper, we study the messy language of ordinary human collaborative conversation, and how a large language model can learn to execute instructions from such conversations.  We isolate several factors that affect this task. 
%In short the agents are engaged in what we shall call a {\em collaborative conversation}. 
 %We use a discourse annotated version of the MDC \cite{narayan:etal:2019,thompson:etal:2024}, the MSDC, in which discourse relations link dialogue moves and action sequences into a coherent narrative structure.  %We then detail how various discourse relations affect the content of dialogue moves and also nonlinguistic action sequences. % -- clean up -- output action sequence to see if the cleaning helps. 

The first factor is the interaction between linguistic and nonlinguistic contexts.  Previous work has shown that at least some context is needed to understand and carry out conversationally given instructions~\cite{jayannavar:etal:2020}.  We improve on that work by first establishing a baseline by using the entire exchange up to an instruction $i$ as a context for an LLM to interpret $i$. Our LLM model, Nebula (\textbf{Ne}ural \textbf{bu}ilder with L\textbf{la}ma), trained on the Minecraft Dialogue Corpus (MDC)~\cite{narayan:etal:2019}, achieves net-action F1 scores that is almost double of~\citet{jayannavar:etal:2020}. Using the Minecraft Structured Dialogue dataset (MSDC)~\cite{thompson:etal:2024}, which provides semantic relations between MDC dialogue moves and nonlinguistic actions, we show that particular discursive components of the linguistic and nonlinguistic context are necessary and sufficient for the LLM to understand an instruction to the degree provided by the baseline.  

Analysing Nebula's output reveals two other factors that adversely affect its performance.  An instruction in the MSDC has two basic components: a description of a shape in terms of four parameters---numbers of components, colors, arrangement and orientation---and the description of a location where the shape should be placed.  Human Architects often use analogies to everyday objects that may be challenging to process; in addition, shape descriptions are often underspecified, meaning that one could perform the instruction correctly in various ways.   Location descriptions in the Minecraft world are also quite difficult to process and highly underspecified. For example, {\em put a tower in a corner} could be correctly located in any of the four corners of the Minecraft board.  We address this problem in two ways: first by further finetuning Nebula on a synthetic dataset to improve its performance in building basic shapes and locating them appropriately; and secondly, and more importantly, by revising the evaluation metric used by~\citet{jayannavar:etal:2020} to reflect more realistically the semantics of location expressions.  We show that, on our synthetic dataset, Nebula achieves high accuracy as per our intuitive metric in performing basic instructions.

The main contributions of this work are: (i) to show that finetuning Llama-3-8B \cite{dubey2024llama} on the entire prior history of the action-prediction task Nebula doubles the net-action F1 score on the Minecraft dataset as compared with the Neural Builder of~\citet{jayannavar:etal:2020}; (ii) to show that training on \textit{Narrative arcs} achieves comparable results with baseline Nebula, and that training on narrative arcs performs better than the instruction-action-instruction input template used by~\citet{jayannavar:etal:2020}; (iii) to highlight the drawbacks of the evaluation metric (i.e., net-action F1), and propose and test a new metric on our synthetic datasets. 
%In this paper we argue that a grasp of the discourse structure in the give and take of a cooperative joint enterprise can make a big difference in communicating effective instructions and carrying them out.  To defend this claim, we investigate several hypotheses.  A first hypothesis is that we can improve the learning of the building agent by concentrating first on instruction to action sequences where Architect and Builder mutually understand each other, and the Builder correctly executes what the Architect intended with his instructions.  We will show how discourse structure enables us to find these exchanges. 

%A more ambitious and more interesting hypothesis is that a discourse structure aware Builder can take advantage of the structure and content of conversational moves to learn more efficiently exploiting the discourse structure to find instructions and actions that reflect amendments or editing due to dialogue moves like clarification questions, corrections to instructions or to action sequences or to both. Other dialogue moves, like ones that elaborate on a previous move or explain a previous instruction, will also help refine instructions that a Builder receives.  The Builder's grasp of these semantic relations can vastly improve the map from instructions to actions to be performed, in cases where Builder and Architect diverge in their understanding.  

After some preliminaries and discussion of prior work (Section~\ref{sec:prelim}), we present our model, Nebula, and its baseline performance in Section~\ref{sec:Nebula}, and then a necessary and sufficient discourse feature to get scores equivalent to the baseline in Section~\ref{sec:discourse}.  In Section~\ref{sec:metric-issues}, we explain several issues associated with the Minecraft corpus. We try to address these issues in Section~\ref{sec:shapes}, where we explain our evaluation metric for underspecified instructions, as well as experiments on our synthetic datasets. We conclude in Section~\ref{sec:conclusion}.

\section{Related Work} \label{sec:prelim}

{\bf MDC}
\citet{narayan:etal:2019} construct a corpus of two person dialogues situated in a simulated Minecraft environment.  The dialogues record conversations about collaborative tasks, in which an Architect and a Builder cooperate to build sometimes complex 3-dimensional shapes out of blocks of six different colors.  The Architect provides instructions, while the Builder is tasked with translating these instructions into actions.  The Builder sometimes asks questions, and the Architect may correct themselves or the Builder, or both, concerning both linguistic and nonlinguistic moves.  The corpus accurately reflects the variety and complexity of actual cooperative conversation. Details on the MDC are in Table \ref{table:corpus-characteristics}. The corpus has also led to related work on conversational interactive agents \cite{kiseleva:etal:2022,mohanty:etal:2023,madge2024large}.

\begin{figure}
    \centering
    \includegraphics[width=0.45\textwidth]{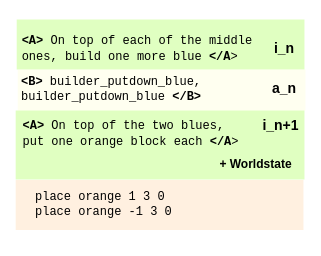}
    \caption{The Neural Builder~\cite{jayannavar:etal:2020} takes as input the sequence $i\_n$ $a\_n$ $i\_n+1$ and the worldstate to predict the subsequent action sequence.}
    \label{fig:neural-input}
\end{figure}

{\bf Instructions to code: Neural Builder and variants} 
The MDC~\cite{jayannavar:etal:2020} incentivizes the development of an algorithm that can predict sequences of actions from instructions.  The actions involved basic moves of placing or removing blocks from certain positions in the environment.~\citet{jayannavar:etal:2020} train a model consisting of a GRU~\cite{cho-etal-2014-properties} to handle textual input coupled with a CNN to integrate information from the current state and a GRU to predicted an action sequence. Although they experiment with several training regimes, the best performance comes from one in which a sequence of conversational moves after some action sequence, assumed to be instructions are given to the model, are followed by the next action sequence of the Builder, followed by the next sequence of linguistic moves are input to the model to predict the subsequent action sequence (See Figure \ref{fig:neural-input}).

% The prediction is then scored relative to how many blocks moved in the predicted action sequence have as a result coordinates that match coordinates of a moved block in the targeted gold action sequence.  In general, they showed that the problem of predicting action sequences from natural language instructions in naturally occurring dialogue remains extremely challenging.  The neural Builder had net action f1 of 0.20 on the MDC test set.

The net-action F1 metric evaluates a model's prediction based on the exact color and coordinate match between the model's predicted sequence and Builder's gold action sequence.  In general,~\citet{jayannavar:etal:2020}  show that the problem of predicting action sequences from natural language instructions in naturally occurring dialogue remains extremely challenging.  Their Neural Builder has net action F1 of $0.20$ on the MDC test set.

\citet{shi:etal:2022} propose a somewhat different task from \citet{jayannavar:etal:2020}; they try to predict when the Builder should execute an action and when they should instead ask for a clarification question.  To this end, they annotate all Builder dialogue moves with a taxonomy of dialogue acts.  They then specify a {\em single} specific action under the execution label instead of a sequence of actions.  Thus, their set-up is not directly comparable to that of \citet{jayannavar:etal:2020}.  

\citet{bonial:etal:2020,bonial:etal:2021} add dialogue acts to Minecraft utterances, but they do not evaluate the effect of these dialogue acts on the Neural Builder's predictions of actions.  Dialogue acts are a partial step towards a full discourse structure: they provide labels for various dialogue moves, but the full discourse structure that we propose to use involves relations between moves.  These relations are important as they tell us how to link different parts of, for instance, an instruction into a coherent whole. As we aim to demonstrate in this paper, discourse structure can help to clean up datasets for training and thereby improve training.  

{\bf MSDC}  \citet{thompson:etal:2024} provide full discourse annotations for the MDC, known as the Minecraft Structured Dialogue Corpus (MSDC), using the discourse theory and annotation principles of SDRT~\cite{asher:1993,asher:lascarides:2003} extended to a multimodal environment, in which both nonlinguistic actions and discourse moves can enter into semantic relations like Elaboration, Correction, and Narration \cite{hunter:etal:2017,asher:etal:2020}.   They follow annotation practices given for the STAC corpus \cite{asher:etal:2016}.  ~\citet{thompson:etal:2024} also adapt the parser from ~\citet{bennis:etal:2023} to predict discourse structures for the MDC with relatively high reliability. Statistics on the MSDC are in Table~\ref{table:corpus-characteristics}.

\begin{table}%[]\
\centering
\begin{tabular}{lrrr}
\toprule
& \multicolumn{1}{c}{\textbf{Train+Val}} & \multicolumn{1}{c}{\textbf{Test}} & \multicolumn{1}{c}{\textbf{Total}}\\ 
\toprule
\multicolumn{4}{l}{\textbf{Original MDC}}\\
\vspace{2\baselineskip}\\
\# Dialogues & 410 & 137 & 547 \\
\midrule
\multicolumn{4}{l}{\textbf{MSDC}} \\
\vspace{2\baselineskip}\\
\# Dialogues & 407 & 133 & 540 \\
\# EDUs & 17135 & 5402 & 22537\\
\# EEUs & 25555 & 7258 & 32813\\
\# EEUs\\ \textit{squished} & 4687& 1473 &6160 \\
\# Relation \\instances & 26279 & 8250 & 34529 \\
\bottomrule
\end{tabular}
 \caption{MDC and MSDC characteristics. EDU and EEU refer to elementary discourse unit, and elementary event units respectively.} 
\label{table:corpus-characteristics}
\end{table}

{\bf LLMs in robotics} Parallel to this work, there has been an increasing amount of research in aiding virtual or real robots with tasks by using LLMs to provide translations from natural language instruction to code that programs the robot to perform the relevant actions \cite{liang:etal:2023,singh:etal:2023,yu:etal:2023}.  This research is directly relevant to our work, as we use LLMs to go from natural language to a pseudo-code of pick and place statements.  However, whereas \citet{liang:etal:2023,singh:etal:2023,yu:etal:2023} focus on optimizing the translation from instructions, typically one instruction, to various different coding paradigms, we focus on how linguistic and nonlinguistic interactions affect the resulting action sequence.  As our results and previous results on the MDC show, producing actions from interactive conversation with frequently underspecified instructions, which are also dependent upon the discourse and nonlinguistic contexts for proper interpretation, is a much more challenging task than translating well crafted unambiguous instructions into code.  In addition, we show that to predict a relevant action from the current instruction $i_{n+1}$ in the MDC environment, it is not sufficient to use a context of just the penultimate instruction $i_n$ and previous action sequence $a_n$.  

\begin{figure}
    \centering
    \includegraphics[width=0.4\textwidth]{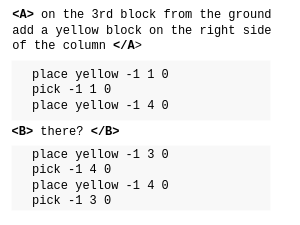}
    \caption{An excerpt from MDC. The Builder interrupts the action sequence by asking a question.}
    \label{fig:minecraft-issues}
\end{figure}

\section{Nebula: an LLM for Predicting Action Sequences} \label{sec:Nebula}

We've seen that \citet{jayannavar:etal:2020}'s Neural Builder performs poorly as per their evaluation method, i.e., net-action F1. The training scheme of \citet{jayannavar:etal:2020} assumes, in effect, that Architect instructions and the Builder actions which fulfill these instructions follow one another in a regular succession. A consequence of this assumption is that actions are individuated by the conversational turns that immediately precede and follow them, and likewise that a new action sequence is initiated whenever there is a linguistic move of any kind. Neural Builder predicts actions from a preceding context containing an instruction, and the instruction just before it, and a representation of the builder moves in between:  $i_{n}a_{n}i_{n+1}$  predicts $a_{n+1}$. Additionally, Neural Builder is provided a worldstate representation (See Figure~\ref{fig:neural-input}).

But this is not realistic, as these bits of text don't always yield a well-formed instruction or even an underspecified one.  In addition, often actions and instructions occur simultaneously.  We might have a clarification question from the Builder in between two action sequences that are in fact carrying out the same action as in Figure~\ref{fig:minecraft-issues}. Builders in the MDC frequently ask questions with respect to the initial instruction about the actions they are currently carrying out; answers to those questions may affect the actions, but it doesn't mean that there are two distinct series of actions pertaining to two distinct instructions, one before the question and its response and one after.  In addition, the Builder sometimes starts to build before the instruction sequence is complete; intuitively, the initial actions form a coherent action sequence with the actions that are subsequent to the further instruction.  These observations show that the assumptions of \citet{jayannavar:etal:2020} about how actions are individuated are too simple.
      
% Different conversational moves will change and make more precise the shape and position of the structure intended by the initial instruction.

The shape and position of the structure, intended by the initial instruction, can change or be made more precise by different conversational moves.~\citet{hunter:etal:2017} note that different conversational moves can help conceptualize actions differently. For example, in many Minecraft sessions, an initial instruction gives the Builder an {\em action type} that might be realized in many different ways.  Something like \emph{build a tower of 5 blocks} is an action type for which a concrete realization would have to specify the color, and perhaps the nature of the building blocks, as well as a location.  As the conversation evolves and unless the Architect corrects their instruction, the type of action to be performed becomes more and more specified.%In other words conversation and conversational structure affects the individuation of actions \cite{hunter:etal:2017}

%As we've already argued, instructions can continue to refine an action that has already been initiated, but at some point we pass onto a new action.

 % One effect of discourse structure on the conceptualization of actions is saying how complex actions get individuated and built up of simpler actions.   How do we decide that stopping point?  One is \cite{jayannavar:etal:2020}'s approach: take all the action sequences that come after a bit of text.  

A simple baseline alternative to the scheme proposed by \citet{jayannavar:etal:2020} that addresses these difficulties is to see how a model performs with the complete prior conversation and action sequences up to the predicted action.  This was not an option for \citet{jayannavar:etal:2020}'s model, but more recent LLMs are capable of doing this.  

We use Llama-2-7B, Llama-2-13B and Llama-3-8B models to take as context all the conversation and action sequences up to action sequence $a_n$ to predict $a_n$.  We finetune Llama on the MDC's~\cite{jayannavar:etal:2020} training set. All the models are finetuned for $3$ epochs using QLoRA method~\cite{qlora}.  Table~\ref{tab:full-run} shows the net-action F1 scores on the validation and test set of MDC. All the finetuned LLMs significantly improve scores in comparison with the $0.20$ F1 score of Neural Builder~\cite{jayannavar:etal:2020}. Llama-3-8B essentially doubles the baseline score of $0.20$. In the rest of this paper, we refer to Llama-3-8B finetuned on MDC as Nebula. The finetuned model, Nebula, and the synthetic datasets used in Section~\ref{sec:shapes} are available here\footnote{\url{https://huggingface.co/linagora/Nebula}}. Table~\ref{tab:model-details} in the Appendix provides details of computing resources and the hyperparameters for finetuning.

\begin{table}[t]
\centering
% \footnotesize
\small
\begin{tabular}{|cccc|}
\hline
Dataset & Llama-2-7b & Llama-2-13b&  Llama-3-8b\\ %Mistral-7b &
\hline
Validation  & 0.292 & 0.323 & 0.398 \\  % 0.286 &
Test  & 0.326 & 0.338 & 0.392  \\ %0.315 &
\hline
\end{tabular}
\caption{Net-Action F1 scores on Minecraft Validation and Test set for predicting action sequences for LLMs using the entire preceding linguistic and non linguistic actions in the game}
\label{tab:full-run}
\end{table}
\noindent

\section{Using Discourse Structure to Improve Nebula}\label{sec:discourse}

The ideal way to model collaborative, instructional interactions like those featured in the MDC is to have two simultaneous, interleaved processes that interact with each other. On one hand, there is the evolving conversational structure that conceptualizes the nonlinguistic actions; on the other, there is the sequence of actions that also affects continuations of the conversational structure. 

This interleaved process is modeled by the discourse structure provided by the MSDC of~\citet{thompson:etal:2024}. The MSDC shows a large scale pattern of \textit{Narrative arcs}. These arcs delimit portions of discourse structure linked by a Narration relation.  Each portion begins with an instruction $i_n$ from the Architect, terminates with an action sequence $a_m$, and involves a negotiation between the Architect and the Builder about the action sequence to be performed. 

The negotiation may be extremely short, where the arc just contains a single instruction and resulting (correct) action $i_n,a_m$.  But it can also contain a complex negotiation involving a number of discourse moves that serve to elaborate or clarify the initial instruction. In this way, the content of a single instruction $i_n$ may evolve over many discourse moves. Further, if the initial builder actions are incorrect, this will prompt corrective moves by the Architect and subsequent revisions by the Builder, leading to longer collaborative negotiations until the Builder carries out the action sequence that satisfies the Architect's initial instruction. 

%connected by relations like Elaboration.  It may also involve Clarification questions or Confirmation questions by the Builder, in which case the instruction {\color{cyan} continues to evolve over many discourse moves} evolves through the portion.  A Narrative arc may also involve actions by the Builder {\color{cyan} that are corrected by and Architect and subsequently revised by the Builder. The action sequence that finally carries out the instruction to the satisfaction of the Architect ends the arc.} %the Architect will correct with a linguistic move that will then result in a nonlinguistic action that revises or corrects the prior actions of the Builder.  The end of the negotiation is the action sequence that finally carries out the instructions to the satisfaction of the Architect.  
\begin{figure}
    \centering
    \includegraphics[width=0.5\textwidth]{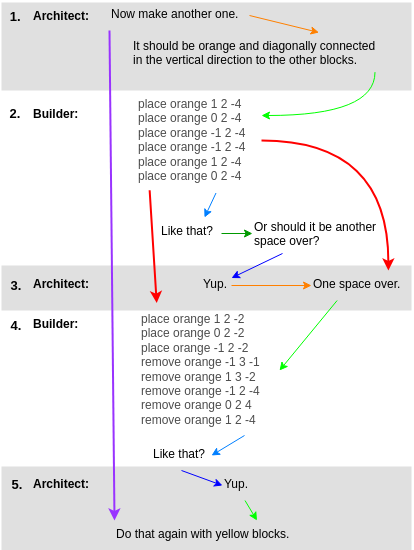}
    \caption{Excerpt of a Narrative arc from the MSDC. Here the arc is purple and connects the instruction in Architect turn one to the following instruction in turn five.}
    \label{fig:narrative-chunk}
\end{figure}

Figure~\ref{fig:narrative-chunk} illustrates a long negotiation enclosed by a Narrative arc. The instruction in the first turn results (in green) in a Builder action sequence in turn two. The Builder then asks a question to confirm that the actions in turn two were correct. The Architect replies to the question by correcting (in red) the actions in turn two, which then results in a corrective action sequence in turn four.

Guided by the assumption that these Narrative arcs contain all the contextual information---the initial instruction and all subsequent discourse moves---that is needed to predict the correct actions, we use the Narrative arcs provided by the MSDC to assign the previous context to each action sequence in the MDC. Since the arcs are automatically recoverable to a relatively high degree by the parsers discussed in \citet{thompson2024llamipa,thompson:etal:2024}, we make the expedient choice to use the gold arcs to test our assumption.

We finetune Llama-3-8B on the MDC training set using instruction-action pairs, where the instruction is only the conversation within the Narrative arc up to the present instruction $i_{n+1}$ in the pair, not the entire conversation history, which we fed to baseline Nebula. Since Neural Builder supplements attenuated contexts with a worldstate representation (see Section~\ref{sec:Nebula}) we also include a ``worldstate'' at the beginning of the Narrative arc in terms of net place actions. We refer to the resulting model as Nebula+N (Nebula trained on Narrative arcs).

%Using the discourse parser {\color{cyan} we used the gold structure?} of~\citet{thompson:etal:2024}, we made a first approximation of these interleaved processes by determining necessary and sufficient situated, conversational conditions for computing instructions. 

% These arcs are automatically recoverable to a relatively high degree by {\color{cyan}methods} discussed in~\citet{thompson:etal:2024} {\color{cyan}and llamipa ref?}
%are relatively self-contained and are recoverable automatically to a relatively high degree by the parser of~\citet{thompson:etal:2024}. 
% So, instead of providing the entire conversation history as in Section~\ref{sec:Nebula} to the model, we provide the worldstate at the beginning of the Narrative arc in terms of net place actions, and the discourse within the Narrative arc up to the present instruction $i_{n}$. We finetune Llama-3-8B on MDC training set using this input {\color{cyan} gold arcs}. We refer to the resulting model as Nebula+N (Nebula trained on Narrative arcs). 

% For discourse parsing, we used \cite{bennis:etal:2023}.    Related work that we plan to use or maybe will use with cleaning instructions involves \cite{bonn2020spatial} AMR annotations and parser based on that. 

% We fine-tuned our LLMs on the 4190 Narration chunks in the MSDC
% training corpus to get the following results on Validation and Test sets for N and N*.

\begin{table}[t] \label{tab:narrative}
\centering
% \footnotesize
\begin{tabular}{|c|c|c|}
\hline
Model  & Validation & Test \\
\hline
%Llama2-13b & 0.323 & 0.338\\
%Llama3-8b & 0.397 & 0.392\\
% Nebula3+N & 0.363 & 0.382\\
Nebula+N & 0.363 & 0.380\\
Nebula+N/N &  & 0.349\\
Nebula+N/$i_{n}a_{n}i_{n+1}$ &  & 0.311\\
 \hline

\end{tabular}
\caption{Net-Action F1 scores on Minecraft validation and test sets for predicting action sequences for LLMs. Nebula+N refers to Nebula trained on Narrative arcs. The next two rows look at those $254$ examples in the test set where $i_{n}a_{n}i_{n+1}$ has less content than the associated Narrative arc. Nebula+N/N gives score of Nebula+N on these samples when worldstate and narrative arc is given as input. Similarly, Nebula+N/$i_{n}a_{n}i_{n+1}$ gives score of Nebula+N on the same samples when worldstate and $i_{n}a_{n}i_{n+1}$ is given as input.}
\label{tab:narrative}
\end{table}

% Table \ref{tab:full-run} gives scores that show that Narrative arcs provided Nebula with sufficient information to approximate the baseline of Table\ref{tab:full-run}. {\color {magenta} can someone put in the details of the experiment?}

Table~\ref{tab:narrative} shows scores on the validation and test sets of the MDC for Nebula+N. We can see that the scores are comparable with Nebula, $0.38$ compared to $0.39$ F1, (see Llama-3-8B in Table~\ref{tab:full-run}).  This provides evidence that the discourse information present in a Narrative arc is \emph{sufficient} for action prediction within that arc. 

The majority of the Narrative arcs are shorter than or of equivalent length to their $i_{n}a_{n}i_{n+1}$ counterparts used to train the Neural Builder ($1321$ out of $1575$ samples in the test set). Going on length alone, it is reasonable to assume Narrative arcs and $i_{n}a_{n}i_{n+1}$ are interchangeable. However, if Narrative arcs are not only sufficient but also \emph{necessary} for action prediction, then in the $254$ samples where Narrative arc is longer than $i_{n}a_{n}i_{n+1}$ we would expect Nebula+N to perform better on those samples when given the Narrative arc. And this is the case. 

In Table~\ref{tab:narrative}, we compare the performance of Nebula+N when worldstate along with the Narrative arc is given as input (denoted as Nebula+N/N), with Nebula+N when worldstate along with $i_{n}a_{n}i_{n+1}$ is given as input (denoted as Nebula+N/$i_{n}a_{n}i_{n+1}$). As we can see, the score for Nebula+N/$i_{n}a_{n}i_{n+1}$ is considerably lower than Nebula+N/N. This provides evidence that Narrative arcs play a crucial role, in that the information they provide is both necessary and sufficient for action prediction in the MDC. 

\section{Problems with the MDC} \label{sec:metric-issues}

In Minecraft, the Architect makes use of several location descriptions. These descriptions are often anaphoric to blocks placed in prior instructions, such as \emph{place another block next to that one} (one that was placed on previous Builder turn); locations are also sometimes vaguely designated (towards the centre) or underspecified (in a corner, along an edge, n blocks/spaces in from an edge/from the centre).  Although the Minecraft environment presents $(x,y,z)$ coordinates, the human participants never used them. This could be because, in the Minecraft environment, players can move their avatars around the board to get different perspectives, which makes it hard to establish an absolute coordinate system.
 
As a result, the net-action F1 metric, which evaluates a model's action sequence based on whether the block placements match exactly in terms of block color and coordinates with the corresponding gold builder action, is often inappropriate. For instance, if the Builder puts down a block at one corner after receiving the instruction {\em in a corner} whereas Nebula chooses another corner, the metric would give Nebula zero credit whereas intuitively it still executed the instruction correctly.  To summarize, the net-action F1 evaluation metric treats vague instructions as completely precise ones, and considers one instantiation of an instruction (i.e. the action sequence of Builder in the gold data) to be the only ground truth. Another related issue is highlighted in~Figure~\ref{fig:minecraft-issues}, where the action sequence for the Architect's instruction gets truncated by a question from the Builder \emph{``there?''}. In this case, for the aforementioned instruction, only the first three actions (\emph{place yellow -1 1 0}, \emph{pick -1 1 0}, \emph{place yellow -1 4 0}) constitute the ground truth.  

Thus, the underspecified instructions with multiple plausible instantiations, coupled with the strict nature of the metric, puts an upper bound on how much the net-action F1 score can improve on this dataset. More importantly, it doesn't reveal what a model with a high F1 score actually does learn. We attempt to answer this in the next section.

% \section{Evaluating and training shapes and locations} \label{sec:shapes}
\section{Evaluating Nebula on Synthetic Data} \label{sec:shapes}

% In this section, we now tackle the two other sources of difficulty for Nebula in mastering instructions: shape and location descriptions. Upon analysing Nebula's performance on the MSDC Builder task, we noticed that the model had recurrent problems with certain shapes, like, square, row, rectangle, tower, diagonal, diamond, cube and location descriptions like centre, edge, corner.  Nebula also failed to understand modifiers like $n\times m$.  

% We also noted that Nebula failed often to place shapes in locations that matched the gold locations that the Builder put them in.

Given the issues associated with MDC and the evaluation metric, we test baseline Nebula (i.e., the one trained on entire conversation history) on simple scenarios using a more just metric. To do so, we construct synthetic datasets at two different levels. The goal of these datasets is to test what basic shapes and location descriptors Nebula learns after being trained on MDC. For the first level, we test Nebula's ability to construct simple shapes, such as, square, row, rectangle, tower, diagonal, diamond, cube of specific size and understand location (i.e. corner, centre, edge)  and orientation descriptions (i.e. horizontal/vertical). We refer to all these shapes as \textbf{level-1 structures}. The resulting dataset, referred to as level-1 dataset, consists of $1368$ instructions. Some of these instructions simply ask to construct a shape of specific size like ``Build a $3\times 3$ red square.'', while others are more detailed, for example, ``Build a $3\times 3$ red horizontal square at the centre.'' 

For rows/diagonals/towers, we vary size from $3$ to $9$. For squares, the size varies from $3\times 3$ to $5\times 5$. For cubes, we only use $3\times 3\times 3$. For rectangles, we use sizes $m\times n$, where $m\neq n$, $m\times n<30$ and $4<=m<=8$.  For diamonds, we use two variants to describe size ``$m$ blocks on a side'' and ``axes $2m+1$ long'', where $3<=m<=6$. We use orientation descriptions (i.e. horizontal/vertical) for squares, rectangles, and diamonds.

\begin{table*}[]
\small
\begin{tabular}{@{}lllllllllllllll@{}}
\toprule
Shape     & \multicolumn{2}{l}{Total \#} & \multicolumn{2}{l}{ShapeAcc\%} & \multicolumn{2}{l}{SizeAcc\%} & \multicolumn{2}{l}{Loc-spec} & \multicolumn{2}{l}{LocAcc\%} & \multicolumn{2}{l}{Orient-spec} & \multicolumn{2}{l}{OrientAcc\%} \\ \midrule
    & \cellcolor{lightgray}BL & FT & \cellcolor{lightgray}BL  & FT & \cellcolor{lightgray}BL & FT  & \cellcolor{lightgray}BL  & FT & \cellcolor{lightgray}BL & FT &\cellcolor{lightgray} BL & FT  & \cellcolor{lightgray}BL  & FT  \\ \midrule
Tower  &  \cellcolor{lightgray}504& 504& \cellcolor{lightgray}100 & 99.0  & \cellcolor{lightgray}100 & 100   & \cellcolor{lightgray}378  &  377&  \cellcolor{lightgray}56.0  &  42.0  &  \cellcolor{lightgray}  &   & \cellcolor{lightgray}  & \\
Row    & \cellcolor{lightgray}168 & 168  & \cellcolor{lightgray}100 &99.0  & \cellcolor{lightgray}100 & 100  &\cellcolor{lightgray}126   &  125 & \cellcolor{lightgray}30.0   & 48.0   &  \cellcolor{lightgray}  &   & \cellcolor{lightgray}  & \\
Diagonal  & \cellcolor{lightgray}168 &168 & \cellcolor{lightgray}78.6& 74.0& \cellcolor{lightgray}95.0 &80.0   & \cellcolor{lightgray}102 & 101  &  \cellcolor{lightgray}2.0  & 39.0  &  \cellcolor{lightgray}  &    & \cellcolor{lightgray}  & \\
Rectangle & \cellcolor{lightgray}140 & 102 &  \cellcolor{lightgray}39.6& 95.0 &\cellcolor{lightgray}12.0 & 49.0 & \cellcolor{lightgray}44 & 76   &  \cellcolor{lightgray}7.0  &  32.0 & \cellcolor{lightgray}31   & 65 & \cellcolor{lightgray}100  & 100  \\
Square   & \cellcolor{lightgray}216 & 144 &  \cellcolor{lightgray}59.3& 89.0 & \cellcolor{lightgray}96.0& 100  &  \cellcolor{lightgray}88 & 93   &  \cellcolor{lightgray}26.0  & 45.0 &  \cellcolor{lightgray}75  & 86   &  \cellcolor{lightgray}81.0 &100  \\
Cube   &  \cellcolor{lightgray}24&24 & \cellcolor{lightgray}58.3& 100 & \cellcolor{lightgray}85.0&100  &  \cellcolor{lightgray}8 & 18 & \cellcolor{lightgray}37.0   & 66.0   &   \cellcolor{lightgray} &  & \cellcolor{lightgray}  &  \\
Diamond  &  \cellcolor{lightgray}144& 108& \cellcolor{lightgray}0 & 18.0 & \cellcolor{lightgray}0& 0  & \cellcolor{lightgray}  &    &   \cellcolor{lightgray} &  &  \cellcolor{lightgray}  & 12   &  \cellcolor{lightgray} & 100 \\
\midrule
Total  & \cellcolor{lightgray}1368 & 1218 & \cellcolor{lightgray}73.0 &87.0  & \cellcolor{lightgray}83.0& 90.0  &  \cellcolor{lightgray}746 & 790 & \cellcolor{lightgray}38.0 & 46.0  &  \cellcolor{lightgray}106  &163    & \cellcolor{lightgray}86.0  & 100  \\
\bottomrule
\end{tabular}
\caption{Evaluation of baseline Nebula (BL) and Nebula finetuned further on a part of synthetic data (FT), on shapes and basic locations. ShapeAcc\% gives percentage of cases where the given shape was correct. Additionally, SizeAcc\% denotes, for the correct shapes, percentage of cases where it is of the correct size; Loc-spec denotes, for the correct shapes, how many have location specified; LocAcc\% denotes location accuracy for such cases. We also test rectangle and square for orientation (horizontal or vertical). Orient-spec denotes, for the correct shapes, the number of cases where orientation is specified; and OrientAcc\% denotes the orientation accuracy for the same.}
\label{tab:shapes:eval}
\end{table*}

\begin{table}[t]
\begin{tabular}{@{}lllll@{}}
\toprule
Instruction          & \multicolumn{2}{l}{Total \#} & \multicolumn{2}{l}{Accuracy(\%)} \\ \midrule
& \cellcolor{lightgray}BL  & FT               & \cellcolor{lightgray}BL           & FT             \\ \midrule
Overall & \cellcolor{lightgray}1368  & 1259        & \cellcolor{lightgray}80.4            & 89.6           \\ \midrule
\multicolumn{5}{l}{Place...}                                                                      \\ \midrule
on top of         & \cellcolor{lightgray}178           & 178          & \cellcolor{lightgray}74.2            & 79.7           \\
to the side of    & \cellcolor{lightgray}154           & 154          & \cellcolor{lightgray}98.1            & 87.7           \\
touching          & \cellcolor{lightgray}176           & 120          & \cellcolor{lightgray}99.4            & 93.3    \\
not touching      & \cellcolor{lightgray}187            &134          & \cellcolor{lightgray}\textbf{7.5}             & \textbf{97.8}           \\
Place Overall        &\cellcolor{lightgray}695           & 586          & \cellcolor{lightgray}67.9            & 88.7           \\ \midrule
\multicolumn{5}{l}{Remove...}                                                                     \\ \midrule
any block         & \cellcolor{lightgray}234           & 234          & \cellcolor{lightgray}95.3            & 94.4           \\
block just placed & \cellcolor{lightgray}216           & 216          & \cellcolor{lightgray}95.3            & 84.7           \\
top block         & \cellcolor{lightgray}44             & 44            & \cellcolor{lightgray}100             & 100            \\
bottom block      & \cellcolor{lightgray}65             & 65            & \cellcolor{lightgray}100             & 100            \\
centre block      & \cellcolor{lightgray}56             & 56            & \cellcolor{lightgray}60.7            & 66.1           \\
corner block      & \cellcolor{lightgray}2               & 2              & \cellcolor{lightgray}100             & 100            \\
end block         & \cellcolor{lightgray}56             & 56            & \cellcolor{lightgray}96.4            & 100            \\
Remove Overall       & \cellcolor{lightgray}673           & 673          & \cellcolor{lightgray}93.3            & 90.3     \\\bottomrule  
\end{tabular}
\caption{Evaluation of baseline Nebula (BL) and Nebula finetuned further on a part of synthetic data (FT), on location descriptors for \emph{place} and \emph{remove} instructions. The FT model performs considerably better for ``not touching'' place instructions while remaining at-par for other instruction types.}
\label{tab:loc:eval}
\end{table}
To evaluate Nebula on these instructions, we use simple binary functions $is\_square(C)$, $is\_tower(C)$ etc. for each shape. These functions take as input the predicted construction $C$ and returns \emph{True} if C is the desired shape, and \emph{False} otherwise. For example, $is\_tower$ checks whether all the blocks have the same value for X and Z (as $Y$ is the vertical dimension) and  Y values are distinct and form a sequence ${1,2, ..., n}$ where $n$ is the number of predicted blocks. 

For an instruction, we first evaluate if the predicted shape is correct. For correct shapes, we evaluate whether the size/color and location/orientation is correct (for instructions where location/orientation was specified). For an instruction with location description like \emph{Build a red tower in a corner}, the location is considered correct if the predicted tower is in any of the four corners.

% So for instance, placing a tower in a corner was counted as correct if Nebula placed a tower in any corner of the board.  

% We evaluated Nebula's understanding  basic shapes with the level 1 dataset. We evaluated whether Nebula could build these shapes of a specified size in any location.  Even without giving a location, shape instructions are often underspecified.  Our location evaluated followed the metric we outlined above.  

% \begin{table*} 
% \centering
% \begin{tabular}{lllllllll}
% \hline
% shape  & corr\# &acc-shape & acc-size & loc-spec & acc-loc & or-spec & acc-or\\
% \hline
% Tower& 504 & 100\%  & 100\%  & 378 & 56.0\% & & \\
% Row& 168 & 100\%  & 100\%  & 126 & 30.0\% & & \\
% diagonal& 168 & 78.6\%  & 95.0\% & 102 & 2.0\%& & \\
% rectangle& 140 & 39.6\%  & 12.0\% & 44 & 7.0\% & 31 & 100\%\\
% square& 216 & 59.3\%  & 96.0\% & 88 & 26.0\% & 75 & 81.0\% \\
% cube& 24 & 58.3\%  & 85.0\% & 8 & 37.0\% & &\\
% diamond& 144 & 0\%  & 0\% &  &  &  & \\
% \hline
% total& 1368 & 73.0\% & 83.0\% &746 & 38.0\% & 106 & 86\%\\
%  \hline

% \end{tabular}
% \caption{Evaluation of Nebula on shapes and basic locations. 
% Shape accuracy gives how many of the shapes Nebula made were correct given the instruction. Additionally: acc-size--of those correct shapes how many were of the correct size; loc-spec--of the correct shapes how many had location specified; loc acc--of those correct shapes with specified locations how many were correctly located. We also tested rectangle and square for orientation (horizontal or vertical).}
% \label{tab:shapes:eval}
% \end{table*}

Table \ref{tab:shapes:eval} gives the result of baseline Nebula (refer to BL in the table) on level-1 dataset. We don't report color accuracy, as Nebula always gets the color correct. From the table, we can see that Nebula already has a decent command of basic shapes like towers, rows, and diagonals. However, it struggles with shapes like rectangle, square, cube, and diamond. It never correctly constructs diamonds, which might be because there are very few instances of diamonds in MDC. For squares and rectangles which are correctly predicted, the model scores very high on orientation accuracy. However, the model has quite low location accuracy across all the correctly predicted shapes. The model rarely achieves an accuracy of above $50\%$, even with our relaxed evaluation method for locations.

We also evaluate the performance of Neural Builder~\cite{jayannavar:etal:2020} on our level-1 dataset. The results are shown in Table~\ref{tab:shapes:eval-nb} of the Appendix. Apart from rows and towers, Neural Builder model fails to build any shape correctly.
 % Such definitions enabled us to formulate better the meanings human agents were assigning to these shapes in Minecraft and to pinpoint where Nebula had trouble.  

As a second step, we look at Nebula's ability to understand location descriptions, in particular ones that are anaphorically specified. To do so, we start with an instantiation (randomly chosen from the set of correct instantiations) for the $1368$ instructions in level-1 dataset. So, for a level-1 instruction such as ``Build a $3\times 3$ red square.'', we have a $3\times 3$ red square already present in the grid.  Now given a level-1 structure in the grid, we design \textbf{level-2 instructions} which require placing or removal of a specific color block. For place instructions, we use location descriptions like \emph{on top of}, \emph{to the side of}, \emph{touching}, and \emph{not touching}. So an example of level-2 place instruction is \emph{``place a blue block on top of that.''} where \emph{that} refers to the level-1 structure in the grid. Similarly, for removal instructions, we have the simple instruction ``remove a block'' and more complex instructions including location descriptions like \emph{you just placed}. We also have additional location descriptions for certain level-1 structures such as \emph{end} for rows, diagonals; \emph{top, bottom} for towers; \emph{corner} for cube; \emph{centre} for cube, odd-size squares and towers. An example of level-2 remove instruction is \emph{``remove the top block.''} Both level-1 and level-2 datasets were generated in an automated fashion.

Similar to level-1, we evaluate Nebula on level-2 dataset by making use of binary functions like $is\_ontopof(b,C)$, $is\_touching(b,C)$ where $C$ is the level-1 structure already present in the grid and $b$ is the predicted block. For example, binary function for \emph{on top of} checks whether there is no block in C which is directly above the block b, and there is a block in C underneath block b.

Table \ref{tab:loc:eval} shows that baseline Nebula performs quite well (refer to BL in the table), with the exception of the instruction involving \emph{not touching} as location description. These results indicate that Nebula has a good knowledge of basic anaphoric location descriptions. Similar to level-1 dataset, we look at the performance of Neural Builder~\cite{jayannavar:etal:2020} on the level-2 dataset.  Table~\ref{tab:loc:eval-nb} in the Appendix shows that the performance of the Neural Builder is considerably worse than Nebula for all the location descriptors.  

We then examine Nebula's errors with \emph{on top of}. We find that the failure cases mostly were a result of the model placing multiple blocks instead of just one on the given level-1 structure. That is, the model does not always understand {\em a block} as {\em a single block.}  In light of these cases, when we check whether all the blocks in predicted $b$ are \emph{on top of} $C$, the accuracy improves from $74.2\%$ to $97.2\%$. Thus, some of the difficulties Nebula had with instructions come from what might be a limited understanding of the semantics and pragmatics of indefinite and numerical noun phrases. We provide visual comparisons of Neural Builder and baseline Nebula on level-1 and level-2 dataset examples in Tables~\ref{tab:lvl1-vis} and~\ref{tab:lvl2-vis} of the Appendix.
% \begin{table}[h] 
% \small
% \flushleft
% \begin{tabular}{cccc}
% \hline
% Instruction  & Accuracy & \#correct & \#total\\
% \hline
% Overall & 0.834& 1141 & 1368\\
% Overall place& 0.738 & 513 &695\\
% Place on top of & 0.972& 173 & 178\\
% Place to the side of& 0.981& 151& 154\\
% Place touching & 0.994 & 175 & 176\\
% Place not touching & 0.075 &14 &187\\
%  \hline
% \end{tabular}
% \caption{Evaluation of place and remove instructions with anaphoric locations using ''all blocks criterion".}
% \label{tab:loc:eval1}
% \end{table}

 \subsection{Finetuning Nebula on Shapes and Locations} 

% \begin{table*} 
% \centering

% \begin{tabular}{lllllllll}
% \hline
% shape  & tot\# &acc-shape & acc-size & loc-spec & acc-loc & or-spec & acc-or\\
% \hline
% Tower& 504 & 99.0\%  & 100\%  & 377 & 42.0\% & & \\
% Row& 168 & 99.0\%  & 100\%  & 125 & 48.0\%& & \\
% diagonal& 168 & 74.0\%  & 80.0\% & 101 & 39.0\% & \\
% rectangle& 102 & 95.0\%  & 49.0\% & 76 & 32.0\% & 65 & 100\%\\
% square& 144 & 89.0\%  & 100\% & 93 & 45.0\% & 86 & 100\% \\
% cube& 24 & 100\%  & 100\% & 18 & 66.0\% &  &\\
% diamond& 108 & 18.0\%  & 0\% & & & 12 & 100\%\\
% \hline
% total& 1218 & 87.0\% & 90.0\% &715 & 46.0\% & 163 & 100\%\\
%  \hline
% \end{tabular}
% \caption{Evaluation of Nebula after finetuning on shapes and basic locations. Shape accuracy (acc-shape), acc-size, location-spec, acc-loc, or-spec and acc-or as in Table \ref{tab:shapes:eval}}
% \label{tab:shapes:fine-tune}
% \end{table*}

Our evaluation on level-1 and level-2 data shows that Nebula struggles with squares, rectangles, diamonds, and ``not touching'' place instructions. To tackle this, we use a subset of the two datasets to augment the training data for Nebula. From level-1 data, we take the following subset for training: squares of size $3\times 3$, diamonds of size $3$ (or axes $5$ spaces long), and rectangles of sizes $4\times 3$ and $5\times 4$. From level-2 data, we take those “touching/not touching” instances where the level-1 structure is square or rectangle. Out of total $363$ instances for touching/not touching, there are $109$ such instances. We then finetune Nebula by combining the Minecraft training with this subset of level-1 and level-2 data. The rest of the level-1 and level-2 data is used for testing.

% \begin{table}[h] 
% \small
% \flushleft
% \begin{tabular}{cccc}
% \hline
% Instruction  & Accuracy & \#correct & \#total\\
% \hline
% Overall  &89.6\%& 1128 &1259\\
% Overall place  & 88.7\%&  520 & 586\\
% Overall removal & 90.3\% & 608 & 673\\
% Place on top of & 79.7\%& 142 &178\\
% Place to the side of & 87.7\%& 135 &154\\
% Place touching &93.3\%& 112& 120\\
% Place not touching  & 97.8\% & 131 &134\\
% Removal any & 94.4\%& 221 & 234\\
% Removal you just placed & 84.7\%& 183& 216\\
% Removal top& 100\% & 44 & 44\\
% Removal bottom & 100\% & 65 & 65\\
% Removal centre& 66.1\%& 37 &56\\
% Removal corner & 100\% & 2 & 2\\
% Removal end & 100\% & 56 & 56\\
%  \hline
% \end{tabular}
% \caption{Evaluation of Nebula with additional finetuning on touching/not touching.}
% \label{tab:loc:fine-tune}
% \end{table}

Table~\ref{tab:shapes:eval} shows Nebula's performance on the level-1 test set after finetuning (refer to FT in the table). As before, we find that Nebula always got the color correct. From the table, we can see that the shape accuracy improves significantly for squares, rectangles, and diamonds in comparison to baseline Nebula (BL in the table). Although the location accuracy is still low, it improves in comparison with baseline Nebula. Interestingly, we also see that Nebula has perfect shape accuracy on cube, although cube is not part of the training set. Finally, for correctly predicted shapes, Nebula achieves a perfect orientation accuracy. 

Table \ref{tab:loc:eval} shows the results on the level-2 test set for Nebula after finetuning. Here also, we can see that Nebula's accuracy remains very high on almost all of the simple instructions with the anaphoric location descriptions. Furthermore, its accuracy increases drastically for ``not touching'' instructions. This jump in accuracy is significant enough to conclude that Nebula has learned the concept of ``contact'', at least for our synthetic dataset. On the minecraft test set, we find that Nebula's performance remains high with a net action F1 of $0.391$. As we can see, these scores are at-par with the baseline Nebula (refer to llama-3-8b in Table~\ref{tab:full-run}).
 
% \subsection{Discussion of evaluation metrics}

% Our interpretation for underspecified location descriptions provides in principle a much more generous evaluation for Nebula's instruction following.  Even our synthetic dataset, using the strict evaluation of matching $x,y,z$ coordinates would produce very low scores for location accuracy given our underspecified instructions, which mirror those of the MDC. Failure to identify the gold Builder placements from instructions accounted for the majority of Nebula's (and NeuralBuilder's) errors on the Minecraft task. 

% Our interpretation is also much truer to the actual semantics of the location descriptions used in MDC and in many cooperative conversational tasks.  After studying the data in the MDC, it is clear that exact locations sometimes, in fact often, don't matter.  It's the {\em relative} locations of the blocks to each other that matter.  Our evaluation matric captures this notion of relative location but leaves exact locations typically underspecified.  This matches our semantic intuitions.

% Our scores on the MDC test section did not go down significantly with additional finetuning on our synthetic dataset for shapes and contact but they did not go up either, because we did not apply our metric to the MDC.  This we leave for future work. 

\section{Conclusions and Future Work}\label{sec:conclusion}
We introduce Nebula, an LLM based action prediction model, for the Minecraft Dialogue Corpus.  As a baseline, Nebula uses the entire Minecraft dialogue up to action $a_n$ to predict $a_n$.  We show that this baseline doubles the net action F1  scores of \citet{jayannavar:etal:2020}.  We then show that certain discourse structures provide necessary and sufficient information for inferring actions at-par with the baseline setting.  We also analyze Nebula's errors on MDC and provide additional finetuning to improve the model's ability to interpret underspecified shape descriptions and anaphorically-specified locations using our synthetic dataset.  This allows us to analyze the shortcomings of the net-action F1 metric, and address them using a more realistic evaluation metric. Our evaluation metric captures the notion of relative location, but leaves exact locations typically underspecified, in accordance with our semantic intuitions. For future work, we plan to explore metrics similar to our relative location metric that can be applied more generally, including on the MDC.  Given the improvement in performance of Nebula after finetuning on our synthetic dataset, we hypothesize that in a more controlled collaborative task, with some pedagogical instructions to the Architect, Nebula could contribute as a useful interface for conversational robots that interact with humans.

\section*{Limitations}

The MSDC contains a great deal of discourse information, including a full discourse structure analysis.  We only use some of this information.  Potentially, we could leverage more information from this dataset to improve Nebula's action prediction performance. We also need to extend our constraints to cover other frequent anaphoric location descriptions in addition to {\em on top of X} and {\em to the side of X}.  Locutions like {\em in front of/ behind, underneath, hanging off, next to}
$(X)$ all have underspecified parameters of either orientation, distance or direction that allow for several correct placements, once $X$ has been identified. We need to evaluate Nebula on these expressions as well. Finally, we need to reevaluate Nebula's predictions as well as the original builder actions in the MDC with our more appropriate metric, which is suited to the underspecified shape and location descriptions used in the corpus.    

% For instance, {\em Under}$(X,Y)$ holds if $Coord_{\{y\}}(X) = \{z-n : y \in Coord_{\{y\}}(Y)\}$ and 
% $Coord_{\{x,z\}}(X) \cap Coord_{\{x,z\}}(Y) \neq \emptyset$.

\section*{Ethics Statement}
Our work here has been to improve the capacities of AI systems in interactive tasks where conversation can be used to optimize performance on collaborative actions.   We see no direct ethical concerns that arise from this work. Though conversationally more capable robots, which could be one downstream application of this work, might require additional conversational strategies as constraints to ensure that participating humans retain the final say with regards to the actions in the collaborative tasks.
%Questions about metrics:
%Our discourse structure isn't incremental.  And we lack symmetry We should in fact be predicting the next disocurse move as well as the next action.   Symmetry and incrementality are lacking.

\section*{Acknowledgements}

For financial support, we thank the National Interdisciplinary Artificial Intelligence Institute ANITI (Artificial and Natural Intelligence Toulouse Institute), funded by the French ‘Investing for the Future–PIA3’ program under the Grant agreement ANR-19-PI3A-000. We also thank the projects COCOBOTS (ANR-21-FAI2-0005) and DISCUTER (ANR-21-ASIA-0005), and the COCOPIL ``Graine'' project of the Région Occitanie of France. This project has also been funded by the France 2030 program and is
funded by the European Union - Next Generation
EU as part of the France Relance. This work was granted access to the HPC resources of CALMIP supercomputing center under the allocation 2016-P23060.  We would also like to thank Bastien Navarri for designing the visualization software used in Tables~\ref{tab:lvl1-vis} and~\ref{tab:lvl2-vis}. 

\clearpage

\appendix
\section{Appendix}\label{sec:Appendix}

\begin{table}[h]
\begin{tabular}{@{}ll@{}}
\toprule
\multicolumn{2}{c}{GPUs}             \\ \midrule
\multicolumn{2}{c}{4 NVIDIA Volta V100} \\ \midrule\midrule
\multicolumn{2}{c}{Hyperparameters} \\ \midrule
Training epochs             & 3     \\
batch size                  & 4     \\
optimizer                   & Adam     \\
learning rate               & 2e-4   \\
\multirow{2}{*}{learning rate scheduler}     & linear warm-up and     \\
                                            & cosine annealing \\
warm-up ratio               & 0.03    \\
gradient clipping           &  0.3    \\
lora r                      & 64     \\
lora (alpha)                & 16     \\
lora dropout ratio          & 0.1     \\
\multirow{2}{*}{lora target modules}         &  Only Attention Blocks\\ 
                                        & (q\_proj, v\_proj)    \\
quantization for Llama-3     & 4-bit NormalFloat \\ \bottomrule
\end{tabular}
\caption{\label{tab:model-details}Details on computing resources and hyperparameters for finetuning Nebula.}
\end{table}

Table~\ref{tab:model-details} gives the hyperparameters used for finetuning Nebula along with the computing resources. We adapt the finetuning code from the following repository\footnote{\url{https://github.com/mlabonne/llm-course/blob/main/Fine_tune_Llama_2_in_Google_Colab.ipynb}}.

% \usepackage{booktabs}

\begin{table*}[]
\begin{tabular}{@{}llllllll@{}}
\toprule
Shape     & Total \# & ShapeAcc\% & SizeAcc\% & Loc-spec & LocAcc\% & Orient-spec & OrientAcc\% \\ \midrule \midrule
Tower  &  504& 75.6 &  20.5 & 303  &  0  &  0  &    0  \\
Row    & 168 & 97.6 & 16.6 & 126   &  32.5   &   0 &   0   \\
Diagonal  & 168 & 0 & 0 &  0& 0  &   0 &   0    \\
Rectangle & 140 & 0 & 0 &  0&  0   &  0  &  0     \\
Square   & 216 &  0 & 0 &  0&  0 &  0   &  0   \\
Cube   &  24&  0 & 0 &   0 &  0 &  0   &    0    \\
Diamond  &  144&  0 &  0 &   0 &  0   &   0  &   0   \\
\midrule
Total  & 1368 &  39.8 & 19.4 &   429 &  9.5 &  0 &   0   \\
\bottomrule
\end{tabular}
\caption{Evaluation of Neural Builder~\cite{jayannavar:etal:2020} on shapes and basic locations. ShapeAcc\% gives percentage of cases where the given shape was correct. Additionally, SizeAcc\% denotes, for the correct shapes, percentage of cases where it was of the correct size; Loc-spec denotes, for the correct shapes, how many had location specified; LocAcc\% denotes location accuracy for such cases. We also tested rectangle and square for orientation (horizontal or vertical). Orient-spec denotes, for the correct shapes, the number of cases where orientation was specified; and OrientAcc\% denotes the orientation accuracy for the same.}
\label{tab:shapes:eval-nb}
\end{table*}

\begin{table}[h]
\begin{tabular}{@{}lll@{}}
\toprule
Instruction          & Total \# & Accuracy(\%) \\ \midrule
         \\ \midrule
Overall & 1368  &  50.9    \\ \midrule
\multicolumn{3}{l}{Place...}                                                                      \\ \midrule
on top of         & 178     & 60.1     \\
to the side of    & 154     & 66.2     \\
touching          & 176     & 98.3     \\
not touching      & 187     & 0.5               \\
Place Overall        & 695             &  55.1  \\ \midrule
\multicolumn{3}{l}{Remove...}                                                                     \\ \midrule
any block         & 234              & 88.5            \\
block just placed & 216         & 14.8             \\
top block         & 44                  & 2.3              \\
bottom block      & 65                    & 35.4              \\
centre block      & 56              & 16.1                \\
corner block      & 2               & 0               \\
end block         & 56                     & 73.2                   \\
Remove Overall       & 673                 & 46.5            \\\bottomrule  
\end{tabular}
\caption{Evaluation of Neural Builder~\cite{jayannavar:etal:2020} on location descriptors for \emph{place} and \emph{remove} instructions. }
\label{tab:loc:eval-nb}
\end{table}

\begin{table*}[htbp]
  \begin{center}
  \begin{adjustbox}{width=1\linewidth}
  \begin{tabular}{ c | l | l }
    \hline
     \bf Lvl-1 Instruction & \bf Neural Builder & \bf Nebula \\
     \hline 
     Build a purple tower of size 3 at the centre.  &
     \rule{0pt}{10ex}
     \begin{minipage}{.21\textwidth}
      \includegraphics[width=\linewidth]{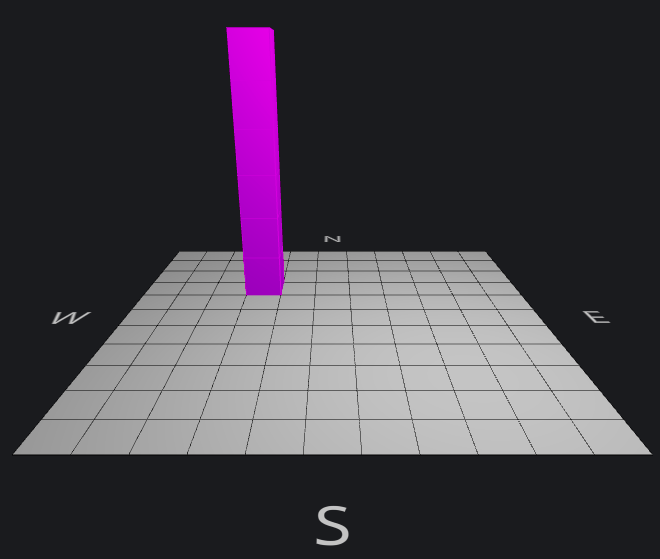}
    \end{minipage}
    & \rule{0pt}{10ex}
     \begin{minipage}{.29\textwidth}
      \includegraphics[width=\linewidth]{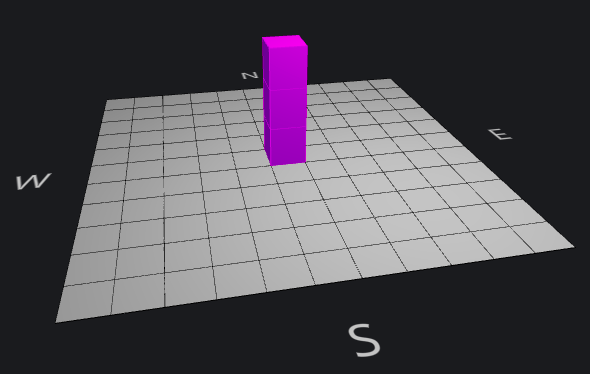}
    \end{minipage}
    \\ 
    [36pt]
    \hline
     Build a 3x3 purple square at the centre.  &
     \rule{0pt}{10ex}
     \begin{minipage}{.2\textwidth}
      \includegraphics[width=\linewidth]{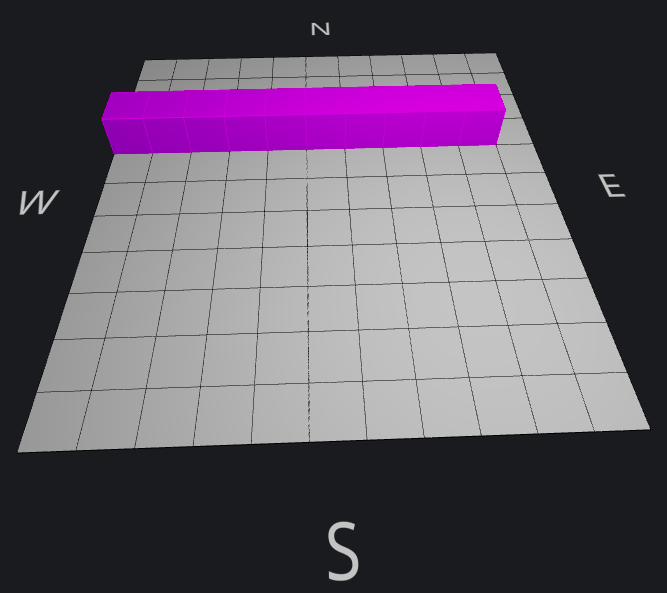}
    \end{minipage}
    & \rule{0pt}{10ex}
     \begin{minipage}{.18\textwidth}
      \includegraphics[width=\linewidth]{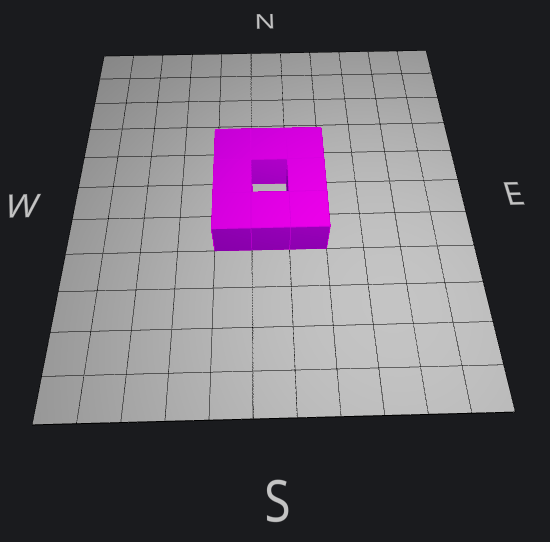}
    \end{minipage}
    \\ 
    [36pt]
    \hline
     Build a diagonal of 3 purple blocks at an edge.  &
     \rule{0pt}{10ex}
     \begin{minipage}{.2\textwidth}
      \includegraphics[width=\linewidth]{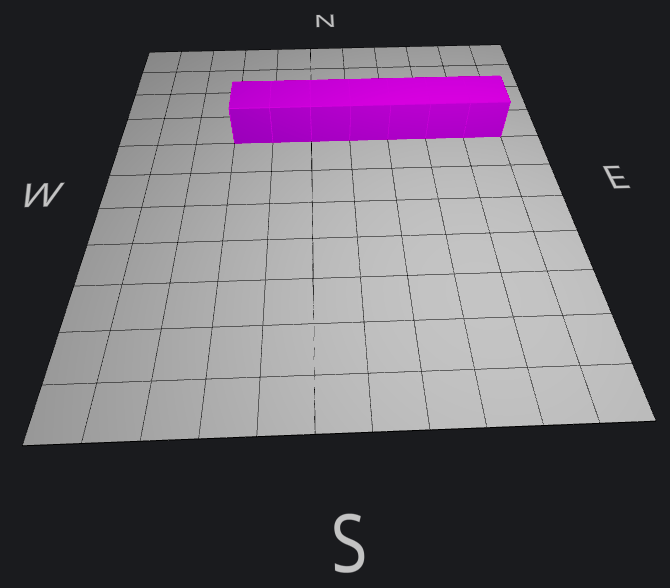}
    \end{minipage}
    & \rule{0pt}{10ex}
     \begin{minipage}{.17\textwidth}
      \includegraphics[width=\linewidth]{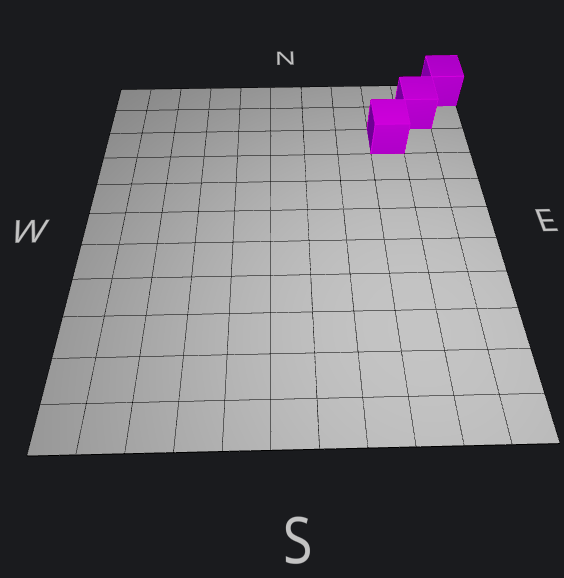}
    \end{minipage}
    \\ 
    [36pt]
    \hline
     Build a 6x3 green rectangle at the centre.  &
     \rule{0pt}{10ex}
     \begin{minipage}{.2\textwidth}
      \includegraphics[width=\linewidth]{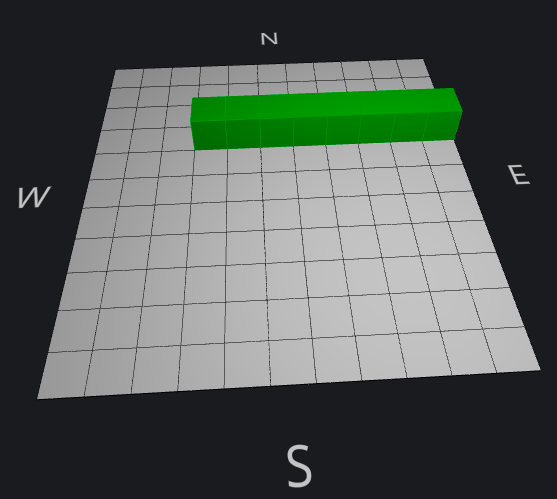}
    \end{minipage}
    & \rule{0pt}{10ex}
     \begin{minipage}{.18\textwidth}
      \includegraphics[width=\linewidth]{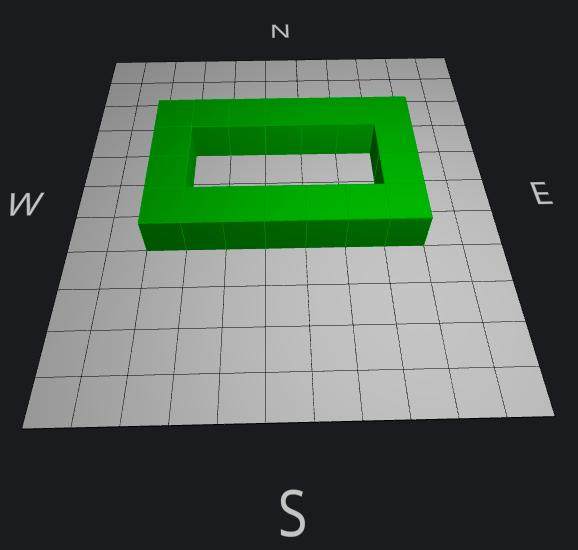}
    \end{minipage} \\ [36pt] \hline
  \end{tabular}
  \end{adjustbox}
  \end{center}
  \caption{Comparison of Neural Builder~\cite{jayannavar:etal:2020} and baseline Nebula on level-1 dataset.}
  \label{tab:lvl1-vis}
\end{table*}

\begin{table*}[htbp]
  \begin{center}
  \begin{adjustbox}{width=1\linewidth}
  \begin{tabular}{ l | l | l}
    \hline
     \bf World State + Lvl-2 Instruction & \bf Neural Builder & \bf Nebula \\
     \hline 
     \rule{0pt}{10ex}
     \begin{minipage}{.3\textwidth}
      \includegraphics[width=0.75\linewidth]{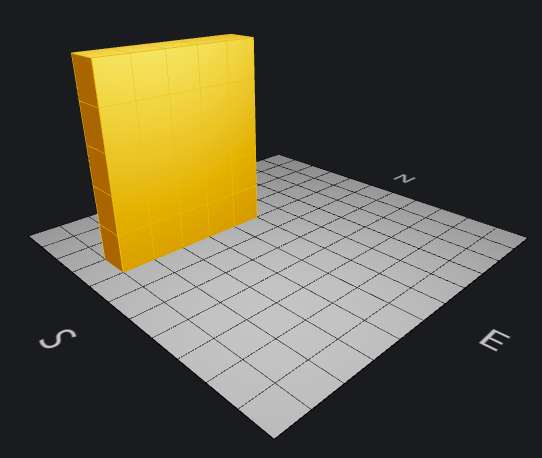} 
    \end{minipage}
    OK now remove the centre block.
    & \rule{0pt}{10ex}
     \begin{minipage}{.22\textwidth}
      \includegraphics[width=\linewidth]{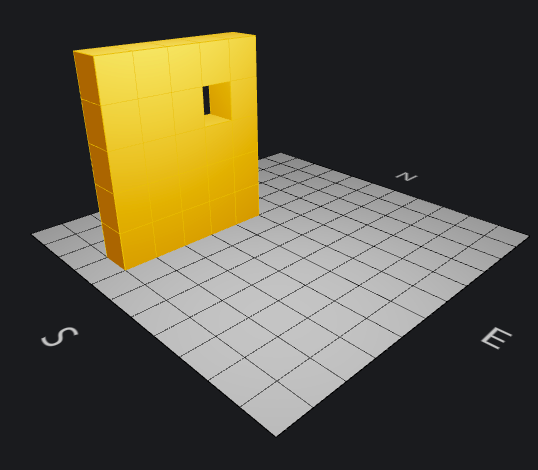}
    \end{minipage}
    & 
    \rule{0pt}{10ex}
     \begin{minipage}{.22\textwidth}
      \includegraphics[width=\linewidth]{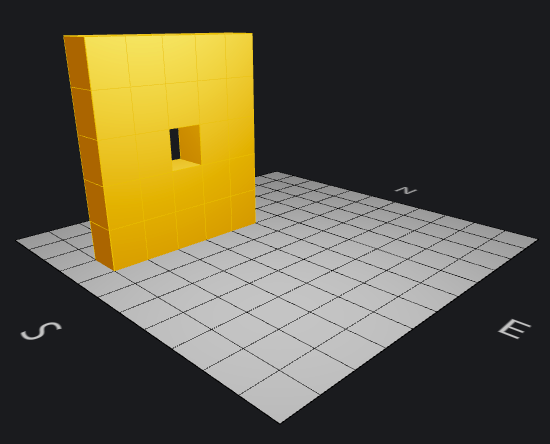}
    \end{minipage}
    \\ 
    [40pt]
    \hline 
       
     \rule{0pt}{10ex}
     \begin{minipage}{.22\textwidth}
      \includegraphics[width=\linewidth]{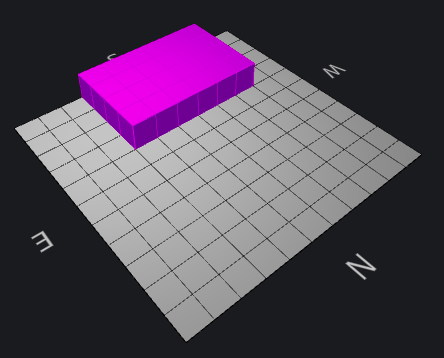}    
    \end{minipage}
    OK now place a yellow block to the side of that.
    & \rule{0pt}{10ex}
     \begin{minipage}{.22\textwidth}
      \includegraphics[width=\linewidth]{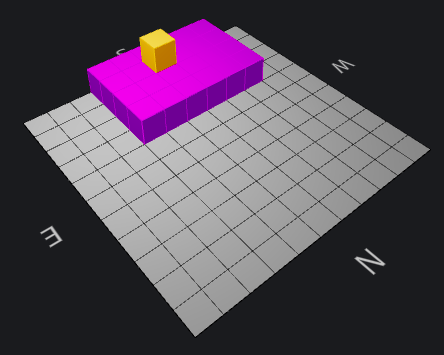}
    \end{minipage}
    & 
    \rule{0pt}{10ex}
     \begin{minipage}{.22\textwidth}
      \includegraphics[width=\linewidth]{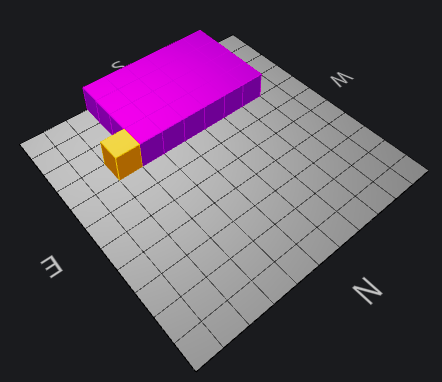}
    \end{minipage}
    \\ 
    [42pt]
    \hline 
       
     \rule{0pt}{10ex}
     \begin{minipage}{.27\textwidth}
      \includegraphics[width=\linewidth]{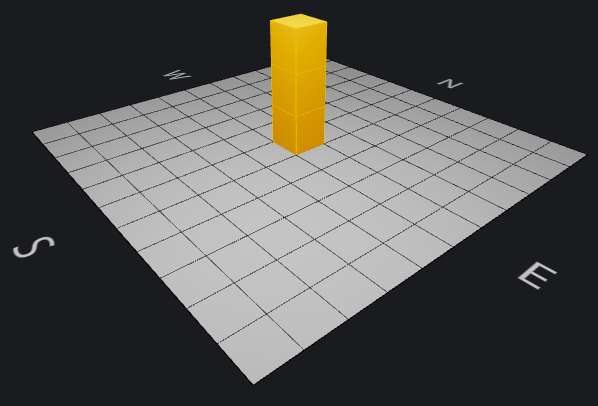}
    \end{minipage}
    OK now remove the top block.
    & \rule{0pt}{10ex}
     \begin{minipage}{.27\textwidth}
      \includegraphics[width=\linewidth]{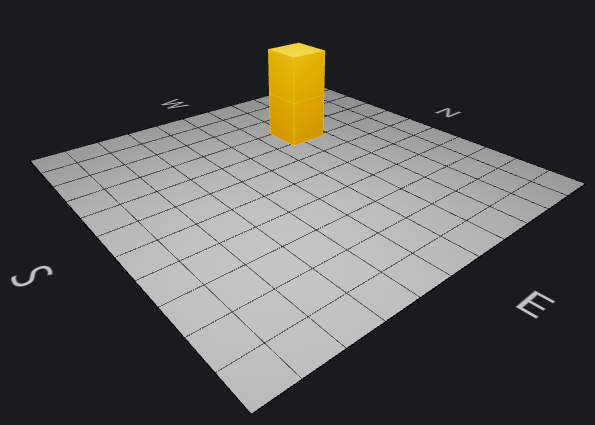}
    \end{minipage}
    & 
    \rule{0pt}{10ex}
     \begin{minipage}{.27\textwidth}
      \includegraphics[width=\linewidth]{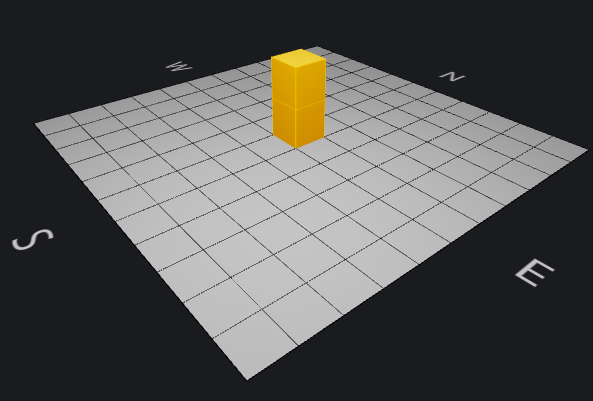}
    \end{minipage}
    \\ 
    [40pt]
    \hline 
     
     \rule{0pt}{10ex}
     \begin{minipage}{.3\textwidth}
      \includegraphics[width=\linewidth]{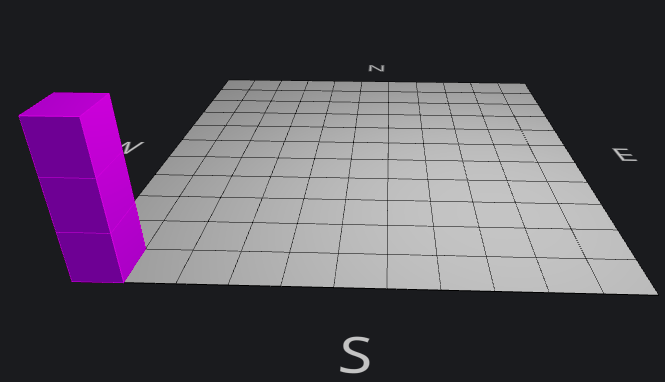}
    \end{minipage}
    OK now remove the bottom block.
    & \rule{0pt}{10ex}
     \begin{minipage}{.3\textwidth}
      \includegraphics[width=\linewidth]{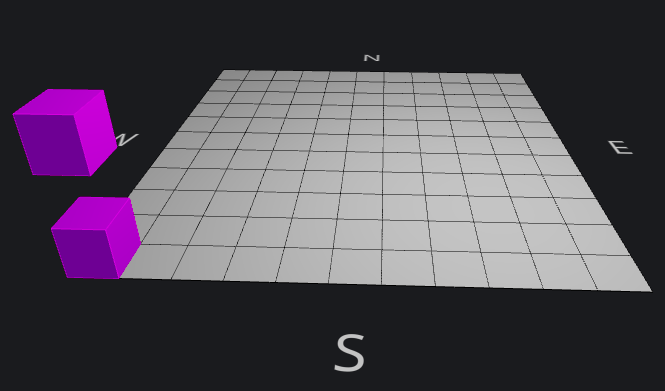}
    \end{minipage}
    & 
    \rule{0pt}{10ex}
     \begin{minipage}{.3\textwidth}
      \includegraphics[width=\linewidth]{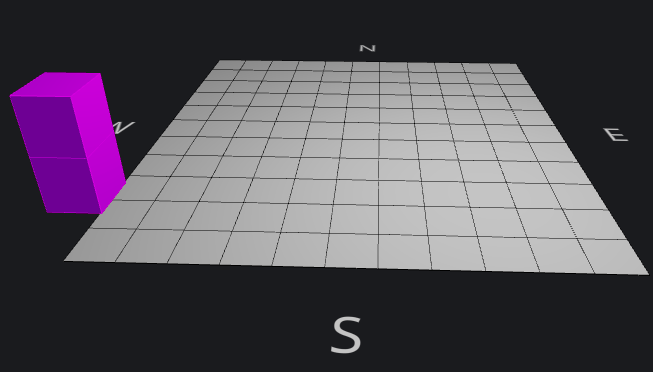}
    \end{minipage}
    \\ [36pt]
    \hline 
  \end{tabular}
  \end{adjustbox}
  \end{center}
  \caption{Comparison of Neural Builder~\cite{jayannavar:etal:2020} and baseline Nebula on level-2 dataset.}
  \label{tab:lvl2-vis}
\end{table*}
\end{document}